\documentclass{article}
\usepackage{arxiv}

\usepackage[utf8]{inputenc} 
\usepackage[T1]{fontenc}    
\usepackage{hyperref}       
\usepackage{url}            
\usepackage{booktabs}       
\usepackage{amsfonts, amsmath}       
\usepackage{nicefrac}       
\usepackage{microtype}      
\usepackage{lipsum}
\usepackage{graphicx}
\usepackage{chngcntr}

\DeclareMathOperator{\Forall}{\forall} 

\makeatletter
\def\blfootnote{\xdef\@thefnmark{}\@footnotetext}
\makeatother

\title{Creating Fair Models of Atherosclerotic Cardiovascular Disease}

\author{
    Stephen Pfohl\thanks{Correspondence to Stephen Pfohl at spfohl@stanford.edu} \\
    \texttt{spfohl@stanford.edu} \\
    Stanford Center for \\
        Biomedical Informatics Research \\
    Stanford University \\
    \and 
    \textbf{Ben Marafino} \thanks{These authors contributed equally} \\
    \texttt{marafino@stanford.edu} \\
    Stanford Center for \\ 
        Biomedical Informatics Research \\
    Stanford University \\
    \\
    \AND 
    Adrien Coulet\footnotemark[2] \\
    \texttt{acoulet@stanford.edu} \\
    Stanford Center for  \\ 
        Biomedical Informatics Research \\
    Stanford University \\
    Universit\'e de Lorraine, \\
        CNRS, Inria, Loria \\
    \and
    \textbf{Fatima Rodriguez} \\
    \texttt{frodrigu@stanford.edu} \\
    Cardiovascular Medicine and \\ 
        Cardiovascular Institute \\
    Stanford University \\
    \AND 
    Latha Palaniappan \\
    \texttt{lathap@stanford.edu} \\
    Primary Care and Population Health \hspace{10cm}\\
    Stanford University \\
    \and
    \textbf{Nigam H. Shah} \\
     \texttt{nigam@stanford.edu} \\
     Stanford Center for \\
        Biomedical Informatics Research \\
     Stanford University \\
}

\begin{document}
\maketitle

\begin{abstract}
Guidelines for the management of atherosclerotic cardiovascular disease (ASCVD) recommend the use of risk stratification models to identify patients most likely to benefit from cholesterol-lowering and other therapies. These models have differential performance across race and gender groups with inconsistent behavior across studies, potentially resulting in an inequitable distribution of beneficial therapy. In this work, we leverage adversarial learning and a large observational cohort extracted from electronic health records (EHRs) to develop a "fair" ASCVD risk prediction model with reduced variability in error rates across groups. We empirically demonstrate that our approach is capable of aligning the distribution of risk predictions conditioned on the outcome across several groups simultaneously for models built from high-dimensional EHR data. We also discuss the relevance of these results in the context of the empirical trade-off between fairness and model performance.
\end{abstract}

\section{Introduction}
Atherosclerotic 
\blfootnote{This work has been accepted to the AAAI/ACM Conference on Artificial Intelligence, Ethics, and Society, 2019. 
This version of the work contains an updated set of experiments based on a refined patient cohort and outcome definition that is more relevant for the intended clinical context. 
In this version, patients are excluded from the cohort on the basis of prior prescription of lipid lowering drugs rather than antihypertensive drugs and fatal coronary heart disease is differentiated from coronary heart disease in the outcome definition.
}
cardiovascular disease (ASCVD), which includes heart attack, stroke, and fatal coronary heart disease, is a major cause of mortality and morbidity worldwide, as well as in the U.S., where it contributes to 1 in 3 of all deaths--many of which are preventable \cite{Benjamin2018}. In deciding whether to prescribe cholesterol-lowering therapies to prevent ASCVD, physicians are often guided by risk estimates yielded by the \textit{Pooled Cohort Equations} (PCEs). The PCEs provide a proportional hazards model \cite{Goff2014} that leverages nine clinical measurements to predict the 10-year risk of a first ASCVD event. However this model has been found to overestimate risk for female patients \cite{Mora2018}, Chinese patients \cite{DeFilippis2017} or globally \cite{Yadlowsky2018}, as well as also underestimate risk for other groups such as Korean women \cite{Jung2015}. Such mis-estimation results in an inequitable distribution of the benefits and harms of ASCVD risk scoring, because  incorrect risk estimates can expose patients to substantial harm through both under- or over-treatment; potentially leading to preventable cardiovascular events or side effects from unnecessary therapy, respectively.

The inability of the PCEs to generalize to diverse cohorts likely owes to both under-representation of minority populations in the cohorts used to develop the PCEs and shifts in medical practice and lifestyle patterns in the decades since data collection for those cohorts. In attempting to correct for these patterns, one recent study \cite{Yadlowsky2018} updated the PCEs using data from contemporary cohorts and demonstrated that doing so reduced the number of minority patients incorrectly misclassified as being high or low risk. Similar results were observed in the same study with an approach using an elastic net classifier, rather than a proportional hazards model. However, neither approach is able to explicitly guarantee an equitable distribution of mis-estimation across relevant subgroups, particularly for race- and gender-based subgroups.

To account for under-represented minorities and to take advantage of the wider variety of variables made available in electronic health records (EHRs), we derive a large and diverse modern cohort from EHRs to learn a prediction model for ASCVD risk. Furthermore, we investigate the extent to which we can encode algorithmic notions of \textit{fairness}, specifically \textit{equality of odds} \cite{Hardt2016}, into the model to encourage an equitable distribution of performance across populations. To the best of our knowledge, our effort is the first to explore the extent to which this formal fairness metric is achievable for risk prediction models built using high-dimensional data from the EHR. We show that while it is feasible to develop models that achieve equality of odds, we emphasize that this process involves trade-offs that must be assessed in a broader social and medical context \cite{Verghese2018}.

\section{Background and Related Work}
\subsection{ASCVD Risk Prediction and EHRs}
The PCEs are based on age, gender, cholesterol levels, blood pressure, and smoking and diabetes status and were developed by pooling data from five large U.S. cohorts \cite{Goff2014} composed of white and black patients, with white patients constituting a majority. Recently, attempts \cite{Yadlowsky2018} were made to update the PCEs to improve model performance for race- and gender-based subgroups using elastic net regression and data from modern prospective cohorts. 
However, this effort focused on demographic groups and variables already used to develop the PCEs and did not consider other populations or clinical measurements. The increasing adoption of EHRs offers opportunities to deploy and refine ASCVD risk models. Efforts have recently been undertaken to apply and refine existing models, including the PCEs and the Framingham score, to large EHR-derived cohorts and characterize their performance in certain subgroups \cite{Pike2016,Rana2016}. Beyond ASCVD risk prediction, there exist many recent works that develop prediction models with EHRs, which are reviewed in \cite{Goldstein2017}.

\subsection{Fair Risk Prediction}
We consider the case where supervised learning is used to estimate a function $f(X)$ that approximates the conditional distribution $p(Y | X)$, given $N$ samples $\{x_i, y_i, z_i\}_{i = 1}^N$ drawn from the distribution $p(X, Y, Z)$. We take $X \in \mathcal{X} = \mathbb{R}^m$ to correspond to a vector representation of the medical history extracted from the EHR prior to a patient-specific index time $t_i$; $Y \in \mathcal{Y} = \{0, 1\}$ to be a binary label, which for patient $i$, indicates the presence of the outcome observed in the EHR in the time frame $[t_i, t_i + w_i]$, where $w_i$ is a parameter specifying the amount of time following the index time used to derive the outcome; and $Z \in \mathcal{Z} = \{0, \ldots, k - 1\}$ indicates a sensitive attribute, such as race, gender, or age, with $k$ groups. The output of the learned function $f(X)$ $\in [0, 1]$ is then thresholded with respect to a value $T$ to yield a prediction $\hat{Y} \in \{0, 1\}$. 

One standard metric for assessing the fairness of a classifier with respect to a sensitive attribute $Z$ is \textit{demographic parity} \cite{Dwork2012}, which evaluates the independence between $Z$ and the prediction $\hat{Y}$. Formally, the demographic parity criterion may be expressed as
\begin{equation} \label{eq:dp_standard}
    p(\hat{Y} | Z = Z_i) = p(\hat{Y} | Z = Z_j) \Forall Z_i, Z_j \in \mathcal{Z}.
\end{equation}
However, optimizing for demographic parity is of limited use for clinical risk prediction, because doing so may preclude the model from considering relevant clinical features associated with the sensitive attribute and the outcome, thus decreasing the performance of the model for all groups \cite{Kleinberg2016}. 

Another related metric is \textit{equality of odds} \cite{Hardt2016}, which stipulates that the prediction $\hat{Y}$ be conditionally independent of $Z$, given the true label $Y$. Formally, satisfying equality of odds implies that
\begin{equation} \label{eq:eo_standard}
    p(\hat{Y} | Z = Z_i, Y = Y_k) = p(\hat{Y} | Z = Z_j,Y = Y_k) \\
        \Forall Z_i, Z_j \in \mathcal{Z}; Y_k \in \mathcal{Y}.
\end{equation}

From this, it can be seen that, if equality of odds is achieved, then for a fixed threshold $T$, both the false positive (FPR) and false negative rates (FNR) are equal across all pairs of groups defined by $Z$. Compared to demographic parity, equality of odds is more appropriate in a clinical setting, since it does not necessarily preclude the learning of the optimal predictor in the case that a true relationship between sensitive attribute and the outcome exists \cite{Hardt2016}.

Furthermore, this definition can be extended to the case of a continuous risk score by requiring that 
\begin{equation} \label{eq:eo_prob}
    p(f(X) | Z = Z_i, Y = Y_k) = p(f(X) | Z = Z_j, Y = Y_k) \\
        \Forall Z_i, Z_j \in \mathcal{Z}; Y_k \in \mathcal{Y}.
\end{equation}
In this case, the distribution of the predicted probability of the outcome conditioned on whether the event occurred or not should be matched across groups of a sensitive attribute. Formulation \ref{eq:eo_prob} is stronger than \ref{eq:eo_standard} since it implies that equality of odds is achieved for all possible thresholds, thus requiring that the same ROC curve be attained for all groups. This is desirable since it provides the end-user the ability to freely adjust the decision threshold of the model without violating equality of odds.

Finally, we also note that satisfying equality of odds for a continuous risk score may be reduced to the problem of minimizing a divergence over each pair $(Z_i, Z_j)$ of distributions referenced in equation (\ref{eq:eo_prob}). \textit{Adversarial learning} procedures \cite{Goodfellow2014} are well-suited to this problem in that they provide a flexible framework for minimizing the divergence over distributions parameterized by neural networks. As such, several related works \cite{Zhang2018,Beutel2017,Edwards2015,Madras2018} have demonstrated the benefit of augmenting a classifier with an adversarial discriminator in order to align the distribution of predictions for satisfying fairness constraints. 

\subsection{Approaches for Achieving Fairness}

Despite considerable interest in the ethical implications of implementing machine learning in healthcare \cite{Char2018,Cohen2014}, relatively little work exists characterizing the extent to which risk prediction models developed with EHR data satisfy formal fairness constraints. 

Adversarial approaches for satisfying fairness constraints (in the form of demographic parity) have been explored in several recent works in non-healthcare domains. One approach, \cite{Edwards2015}, in the context of image anonymization, demonstrated that representations satisfying demographic parity could be learned by augmenting a predictive model with both an autoencoder and an adversarial component.
The adversarial approach to fairness was further investigated by \cite{Beutel2017} with a gradient reversal objective for data that is imbalanced in the distribution of both the outcome and in the sensitive attribute. 

In attempting to address the limitations of demographic parity as a metric, \cite{Hardt2016} introduced equality of odds as an alternative and devised post-processing methods to achieve it for fixed-threshold classifiers. Recently, \cite{Zhang2018} and \cite{Madras2018} generalized the adversarial framework to achieve equality of odds by providing the discriminator access to the value of the outcome. 

Both demographic parity and equality of odds are referred to as \textit{group fairness} metrics since they are concerned with encouraging an invariance of some property of a classifier over groups of a sensitive attribute. While straightforward to compute and reason about, optimizing for these metrics may produce models that are discriminatory over structured subgroups within and across groups of sensitive attributes, constituting a form of fairness gerrymandering \cite{Kearns2018}. The competing notion of \textit{individual fairness} \cite{Dwork2012} and may be able to address these concerns by assessing whether a model produces similar outputs for similar individuals. However, this notion is often of limited practical use due to the challenges of developing a domain-specific similarity metric that encodes desired notions of fairness. 

Recent efforts \cite{hebert2018multicalibration} have investigated an alternative to both group and individual fairness metrics with a process that audits a classifier to discover subgroups for which the model is under-performing and iteratively improve model performance for those groups, ultimately resulting in a non-negative change in model performance for all computationally-identifiable subgroups.

The closest related work examining the fairness of risk prediction models in healthcare is \cite{ChenJohanssonSontag_NIPS18}, which, in the context of mortality prediction in intensive care units, argued that any trade-off between model performance and fairness across subgroups is undesirable. They propose that the prediction error should be decomposed in terms of bias, variance, and noise and that the relative contribution of these terms be used to guide additional data collection. 

\section{Methods}

\begin{table}[tb]
\centering
\caption{Cohort characteristics. The number of patients extracted, the incidence of the ASCVD outcome and the average length of follow-up for each subgroup are shown.} \label{tab:cohort_characteristics}
\begin{tabular}{lrrr}
\toprule
           Group &   Count &  ASCVD Incidence & Follow-up Length (Years) \\
\midrule
           Asian &   34,156 &     0.0144 &      2.83 \\
           Black &    9,018 &     0.0271 &      2.76 \\
 Hispanic &   21,587 &     0.0152 &      2.81 \\
           Other &   19,100 &      0.013 &      2.78 \\
         Unknown &   30,300 &    0.00512 &      2.96 \\ 
           White &  136,348 &     0.0141 &      2.93 \\ \midrule
          Female &  154,266 &     0.0116 &      2.87 \\
            Male &   96,074 &     0.0167 &      2.93 \\ \midrule
         40-55 &  117,510 &    0.00603 &      2.95 \\
         55-65 &   64,477 &     0.0128 &      2.92 \\
         65-75 &   44,149 &       0.02 &      2.83 \\
         75+ &   24,373 &     0.0398 &      2.64 \\ \midrule
             All &  250,509 &     0.0135 &      2.89 \\
\bottomrule
\end{tabular}
\end{table}

\begin{table*}[t]
 \centering
 \caption{Distribution alignment metrics. We report the coefficient of variation (CV; the ratio of the standard deviation to the mean) of the false positive rate (FPR, CV) and false negative rate (FNR, CV) at a fixed decision threshold of 0.075 across the race, gender, and age groups. Furthermore, we compute the pairwise earth mover's distance (EMD) between distributions of the predicted probabilities of having an ASCVD event, conditioned on the true ASCVD label $y$ for each group of each sensitive attribute and take the mean.} \label{tab:dist_summary}
\begin{tabular}{lrrrrrr}
\toprule
{} & \multicolumn{2}{c}{Race} & \multicolumn{2}{c}{Gender} & \multicolumn{2}{c}{Age} \\
\cmidrule(lr){2-3}
\cmidrule(lr){4-5}
\cmidrule(lr){6-7}
{} & Standard & \texttt{EQ\textsubscript{race}} & Standard & \texttt{EQ\textsubscript{gender}} &        Standard & \texttt{EQ\textsubscript{age}} \\
\midrule
FNR, CV  &    0.267 &  0.0437 &    0.118 &   0.0072 &           0.347 &  0.0426 \\
FPR, CV &    0.451 &   0.563 &    0.477 &    0.231 &           0.927 &   0.522 \\
Mean EMD $\mid y = 0$      &  0.00905 & 0.00109 &   0.0107 & 0.000983 &          0.0277 & 0.00162 \\
Mean EMD $\mid y = 1$      &   0.0264 & 0.00934 &   0.0144 &  0.00259 &          0.0363 &  0.0189 \\
\bottomrule
\end{tabular}
\end{table*}
\subsection{The Dataset and Cohort Definition}
We extract records from the Stanford Medicine Research Data Repository \cite{Lowe2009}, a clinical data warehouse containing records on roughly three million patients from Stanford Hospital and Clinics and Lucile Packard Children's Hospital for clinical encounters occurring between 1990 and 2018. We define a prediction task that resembles the setting in which the PCEs were developed \cite{Goff2014} for the purpose of guiding physician decision-making in ASCVD prevention and construct a corresponding cohort. 
As a first step, we identify all patients with at least two clinical encounters over a time span of at least two years for which they are 40 years or older. Then, for each patient, we randomly select a clinical encounter from the time span that allows for at least one year of history and one year of follow-up and set the date of the encounter as the index time $t_i$. Patients with no encounters meeting this criteria are excluded.
We further exclude from the cohort patients with a historical diagnosis of cardiovascular disease (myocardial infarction, stroke, atrial fibrillation, heart failure) or a prescription of a lipid lowering drug (see supplementary material) in the five years prior to the index time. 

We assign a positive ASCVD label if a diagnosis code for an ASCVD event (myocardial infarction, stroke, or fatal coronary heart disease) is observed at any point in their record following the index time. We consider coronary heart disease to be fatal if a diagnosis code for coronary heart disease is followed by death within a year.

The patients are randomly partitioned such that 80\%, 10\%, 10\% are used for training, validation, and testing, respectively. The clinical concepts used to define the exclusion criteria and outcome definition are provided in the supplementary material.

\subsection{Sensitive Attributes}
We consider race, gender, and age as sensitive attributes and assess model performance and fairness with respect to them. For race, we use both race and ethnicity variables to partition the cohort into six disjoint groups: Asian, Black, Hispanic, Other, Unknown, and White. Patients not considered Hispanic thus have either a non-Hispanic or unknown ethnicity. For gender, we partition the cohort into male and female populations. For age, we discretize the age at the index time into four disjoint groups: 40-55, 55-65, 65-75, and 75+ years, where the intervals are inclusive on the lower bound and exclusive on the upper bound. A summary of these groups is presented in Table \ref{tab:cohort_characteristics}.
\subsection{Feature Extraction}
For feature extraction, we adopt a strategy similar to the one described in \cite{Reps2018} to convert time-stamped sequences of clinical concepts across several domains (i.e., diagnoses, procedures, medication orders, lab tests, clinical encounter types, departments, and other observations) into a static representation suitable for modeling. For each extracted patient, we filter the historical record to include only those concepts occurring prior to the index time. We encode as a binary attribute each unique clinical concept observed in the dataset according to whether that concept was present anywhere in the patient's history prior to the prediction time; otherwise, it is absent or missing. Similarly, we do not use the numeric results of lab tests or vital measurements, but only include the presence of their measurement. In all models, we include race, gender, and age as features without regards as to whether the variable is treated as sensitive or not. 

\subsection{Adversarial Learning for Equality of Odds}
To develop an ASCVD risk prediction model that satisfies the definition of equality of odds in (\ref{eq:eo_prob}), we consider two fully-connected neural networks: a classifier $f: \mathbb{R}^m \rightarrow \mathbb{R} \in [0, 1]$ parameterized by $\theta_f$ that predicts the probability of the ASCVD outcome $Y$ given data $X$; and a discriminator $g : \mathbb{R} \times \{0, 1\} \rightarrow [0, 1]^k$ parameterized by $\theta_g$ that takes as input both the logit of the output of $f$ and the value of the true label $Y$ to predict a distribution over the groups of a sensitive attribute $Z$. If $L_{cls}$ and $L_{adv}$ are the cross-entropy losses of the classifier predictions over $Y$ and the discriminator predictions over $Z$, respectively, and $\lambda$ is a positive scalar, then the training procedure may be described by alternating between the steps
\begin{equation}
    \min_{\theta_f} L_{cls} - \lambda L_{adv} \quad \mbox{and} \quad \min_{\theta_g} L_{adv}.
\end{equation}
\subsection{Model Training and Evaluation}
The training procedure is composed of four experiments and thus produces four prediction models. The first model is trained to predict the risk of ASCVD and does not use adversarial training. The other three models result from separate training runs in which each of the discrete race, gender, and age variables are considered as sensitive attributes in the adversarial training procedure. We refer to these four experiments as \texttt{Standard}, \texttt{EQ\textsubscript{race}}, \texttt{EQ\textsubscript{gender}}, and \texttt{EQ\textsubscript{age}}. 

For each model, we compute standard metrics on the entire test set and on each subgroup. Specifically, we report the area under the receiver operating characteristic curve (AUC-ROC), the area under the precision-recall curve (AUC-PRC), the Brier score \cite{Brier1950} as a measure of calibration, and the false positive and false negative rates (FPR, FNR) at a fixed threshold of $T=0.075$, in keeping with current ASCVD guidelines for the prescription of statin therapy \cite{Goff2014,Yadlowsky2018}. To express adherence to the standard equality of odds definition in equation \ref{eq:eo_standard}, we report the coefficient of variation (i.e. the ratio of the standard deviation to the mean) of the FPR and FNR at $T=0.075$ across the groups of each sensitive attribute. To assess the distance between the distributions presented in (\ref{eq:eo_prob}), we compute the earth mover's distance (EMD, or first Wasserstein distance) between the empirical distributions of the predicted probability of ASCVD conditioned on whether ASCVD occurred or not for each group of each sensitive attribute in a pairwise fashion and take the mean within each outcome strata.

For each of the four models, we employ fully-connected feedforwad neural networks composed of layers of fixed size for both the classifier and discriminator. Hyperparameters were selected separately for each model across 100 iterations of random search over a grid that included the number of classifier and discriminator layers, the size of each layer, the learning rates of the Adam optimizers \cite{kingma2014adam} applied on $\theta_f$ and $\theta_g$, the discriminator weight $\lambda$, the use of layer normalization \cite{lei2016layer} in the classifier and the discriminator, and the use of spectral normalization \cite{miyato2018spectral} in the discriminator. Models we trained for 100 epochs, where each epoch is composed of 100 one hundred randomly sampled batches of the data. During training, early stopping was performed on the basis of the maximal AUC-ROC on the validation set. As it is not clear how to best perform model selection for the \texttt{EQ} experiments, we take the model with the highest degree of distribution alignment on the validation set, computed as the mean of the EMD metrics computed on each outcome strata, that also achieves an AUC-ROC of greater than 0.7 on the validation set. All training and model development was performed with the PyTorch library \cite{paszke2017automatic}.

\section{Results}
\begin{figure}
\centering
\includegraphics[width=0.99\linewidth]{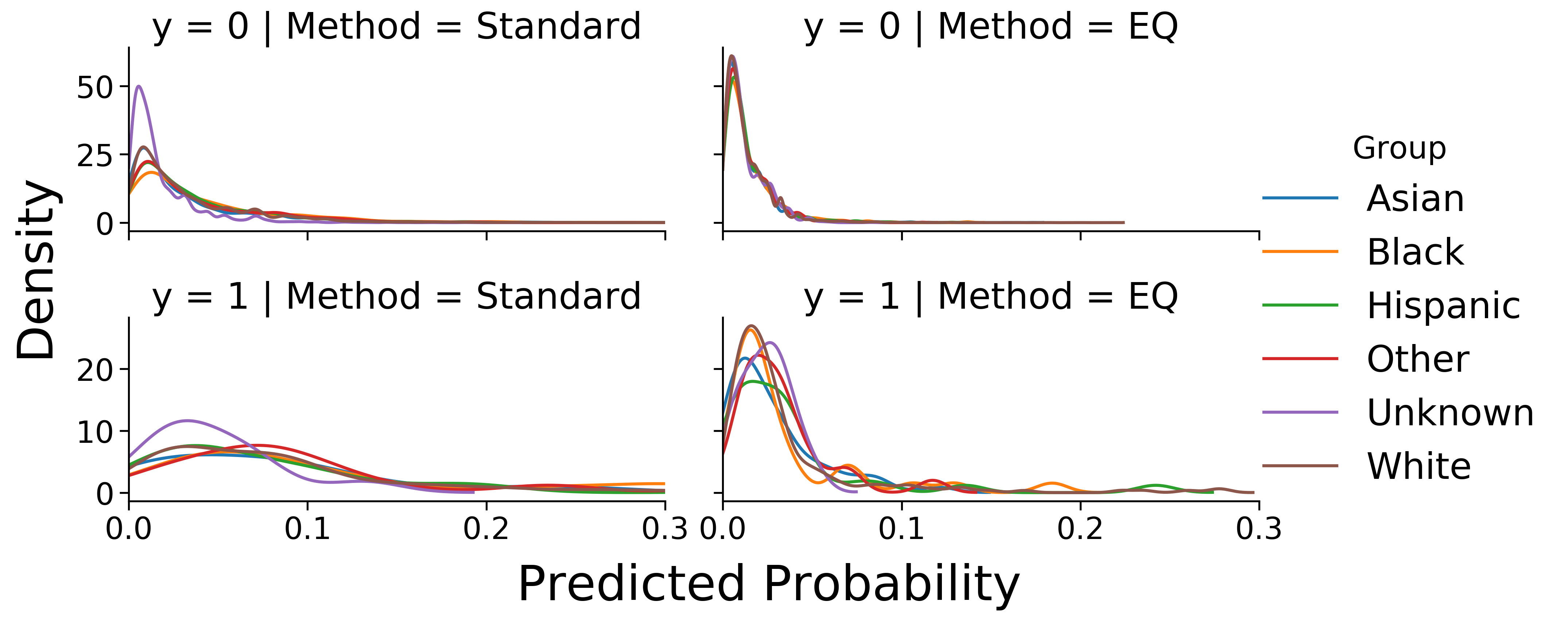} \\
\includegraphics[width=0.99\linewidth]{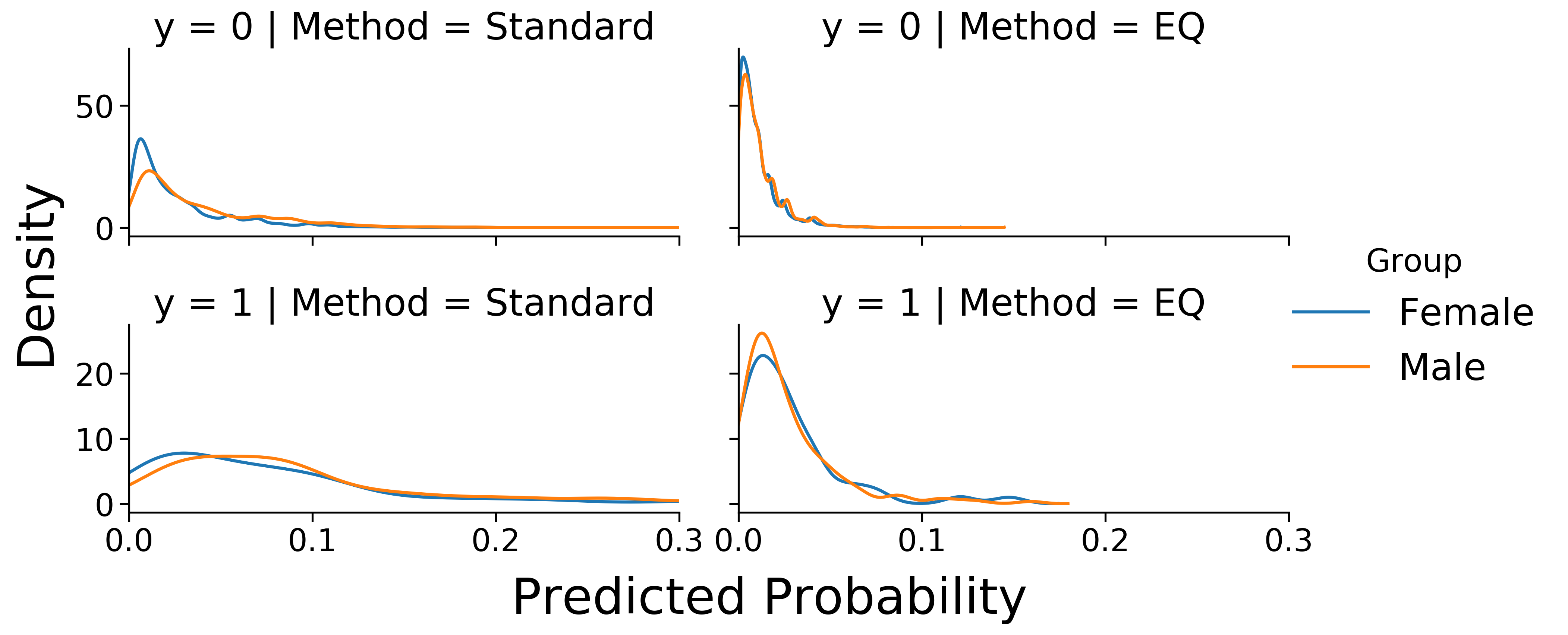} \\
\includegraphics[width=0.99\linewidth]{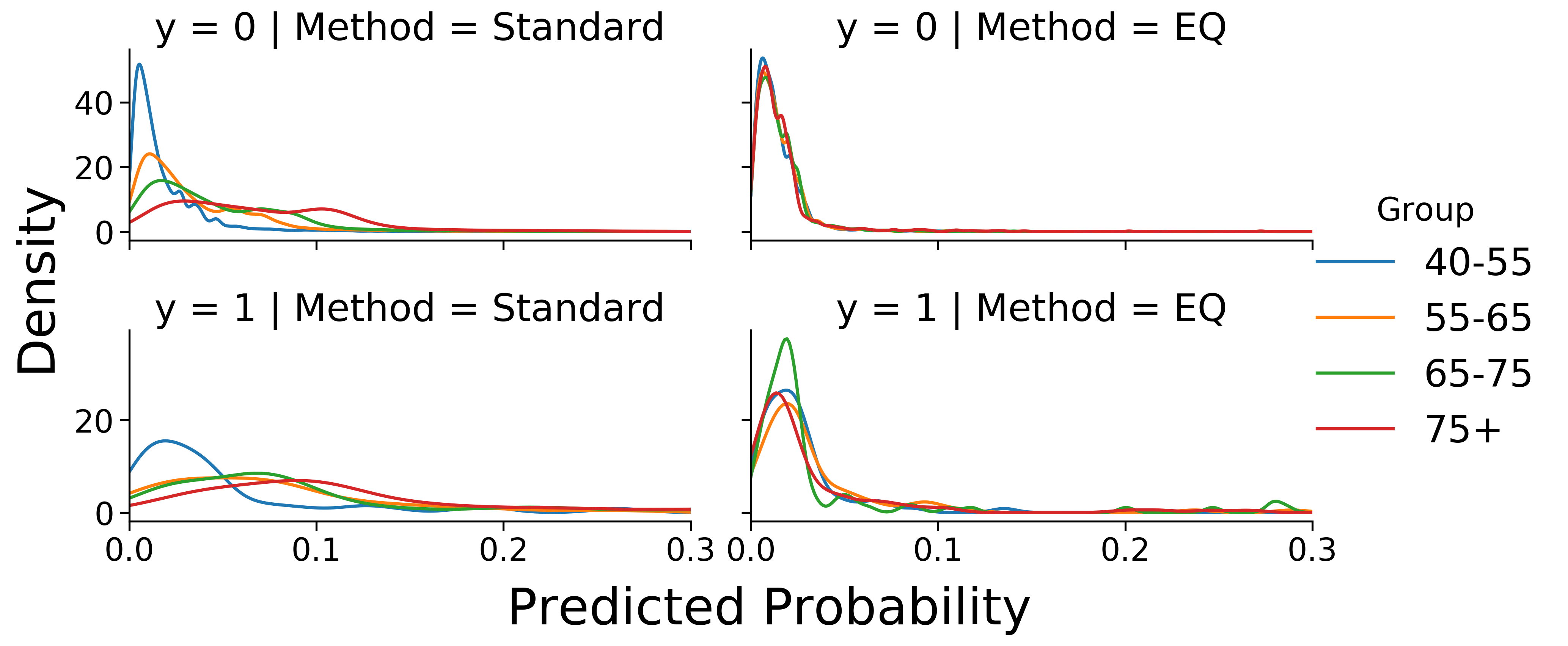}
    \caption{Empirical distribution of the predicted probability of developing ASCVD in the followup period conditioned on whether ASCVD occurred. Plots are stratified by experimental condition (\texttt{Standard} or \texttt{EQ}), true value of the ASCVD outcome ($y = 0$ or $y = 1$), and the variable treated as sensitive (race, gender, or age).}
    \label{fig:prob_dist}
\end{figure}

\begin{table}
\centering
\caption{Model performance measured on the test set without stratification for each experimental condition.} \label{tab:results_whole_pop}
\begin{tabular}{lrrrr}
\toprule
{} &  Standard &  \texttt{EQ\textsubscript{race}} &  \texttt{EQ\textsubscript{gender}} &  \texttt{EQ\textsubscript{age}} \\
\midrule
AUC-ROC   &     0.773 &     0.742 &   0.743 &            0.694 \\
AUC-PRC &    0.0904 &     0.083 &  0.0924 &           0.0887 \\
Brier Score &    0.0143 &    0.0137 &  0.0138 &           0.0136 \\
\bottomrule
\end{tabular}
\end{table}

\begin{table*}
 	\centering
 	\caption{Model performance measured on the test set stratified by group and experimental condition. EQ corresponds to training for the sensitive attribute corresponding to the subgroup of interest.} \label{tab:by_group} 
 \begin{tabular}{lrrrrrr}
\toprule
{} & \multicolumn{2}{c}{AUC-ROC} & \multicolumn{2}{c}{AUC-PRC} & \multicolumn{2}{c}{Brier Score} \\
\cmidrule(lr){2-3}
\cmidrule(lr){4-5}
\cmidrule(lr){6-7}
{} &  \texttt{\texttt{Stand.}} &    \texttt{EQ} &  \texttt{Stand.} &     \texttt{EQ} &  \texttt{Stand.} &      \texttt{EQ} \\
\midrule
Asian           &    0.726 &   0.704 &   0.0775 &  0.0508 &   0.0142 &  0.0133 \\
Black           &    0.785 &    0.74 &    0.244 &   0.158 &   0.0294 &  0.0307 \\
Hispanic &    0.721 &   0.724 &   0.0811 &   0.113 &   0.0157 &  0.0146 \\
Other           &    0.783 &   0.792 &   0.0887 &   0.097 &   0.0133 &  0.0123 \\
Unknown         &    0.824 &    0.77 &   0.0274 &  0.0141 &    0.004 & 0.00376 \\
White           &    0.768 &   0.744 &   0.0956 &  0.0914 &   0.0155 &   0.015 \\ \midrule
Female          &    0.768 &   0.749 &   0.0835 &     0.1 &   0.0123 &  0.0119 \\
Male            &    0.769 &   0.731 &    0.104 &  0.0847 &   0.0175 &  0.0169 \\ \midrule
40-55         &    0.726 &   0.689 &   0.0369 &  0.0398 &  0.00661 &  0.0063 \\
55-65         &    0.741 &   0.742 &   0.0966 &   0.116 &   0.0141 &  0.0131 \\
65-75         &    0.699 &   0.672 &   0.0966 &  0.0919 &   0.0209 &  0.0198 \\
75+         &    0.679 &   0.662 &    0.135 &   0.153 &   0.0408 &  0.0395 \\
\bottomrule
\end{tabular}
\end{table*}

\subsection{Cohort Characteristics}
The cohort extraction procedure produces a cohort of 250,509 patients having 200,407 features, with 3,388 patients labeled as positive for ASCVD (Table \ref{tab:cohort_characteristics}). We note that in this cohort, there are 136,348 white patients, constituting a majority, and 9,018 black patients. Across race groups, ASCVD rates range from 1.0-2.0\%, with the exception of patients with unknown race, who experience a reduced rate of 0.52\%. Furthermore, we observe higher ASCVD rates for male patients compared to female patients. Finally, ASCVD rates appear to increase with age, with rates ranging from approximately 0.6\% for the 40-55 age group to 4\% for patients age 75 or older.
\subsection{Distribution Alignment with Adversarial Training}
Applying the adversarial training procedure results in a alignment of the distributions of the predicted probability of ASCVD conditioned on the true outcome label (Figure \ref{fig:prob_dist}). Without employing a adversarial discriminator, the center of mass of these distributions appears to depend on the base ASCVD rate in the group. However, these differences largely disappear when training in an adversarial setting. This results in a substantial reduction in the mean pairwise EMD between each predictive distribution in both outcome strata for all sensitive attributes (Table \ref{tab:dist_summary}). Furthermore, we note that variability in the FPRs and FNRs at a fixed threshold of $0.075$ is greatly reduced following adversarial training (Table \ref{tab:dist_summary}), indicating that the approach successfully encourages the model predictions to satisfy equality of odds. 

The relative lack of success in minimizing the mean pairwise EMD between the conditional predictive distributions across racial groups that experience ASCVD (Table \ref{tab:dist_summary}) may be largely explained by the anomalous characteristics of the group of patients having unknown race. For instance, when using standard training (\texttt{Standard}), the predictive distribution for both outcome strata for the unknown race group is clearly separated from that of the five groups while the distributions for those five are mostly aligned (Figure \ref{fig:prob_dist}). However, when training the model in an adversarial setting, it appears that the primary effect is to align the predictive distribution for the unknown race group to the region inhabited by the distributions of the remaining groups while disturbing the relative alignment between the distributions for those groups. 
\subsection{The Cost of Fairness}
Satisfying equality of odds with an adversarial objective incurs a reduction in AUC-ROC, AUC-PRC, and calibration for the population at large (Table \ref{tab:results_whole_pop}), with the largest negative effects observed when training to adjust for the differences across age groups (\texttt{Standard} AUC-ROC = 0.773 vs. \texttt{EQ\textsubscript{age}} AUC-ROC = 0.694). 
This procedure induces a reduction in AUC-ROC and AUC-PRC for many of the populations assessed (Table \ref{tab:by_group}), with positive effects observed only in the Hispanic and Other race groups.
Furthermore, while exactly satisfying equality of odds implies that the ROC curve be the same across for each group of a sensitive attribute, the procedure we apply does not empirically produce a model that achieves the same AUC-ROC on each group.

It has been shown that developing a well-calibrated model is an objective that conflicts with that of satisfying equality of odds \cite{Pleiss2017,Kleinberg2016}. In our case, we do not observe a trade-off, as the Brier score is reduced, indicating better calibration, when training for equalized odds both for the aggregate population (\ref{tab:results_whole_pop}) and for many subgroups (Table \ref{tab:by_group}).

\section{Discussion}
We have demonstrated the capabilities of adversarial training procedures to encourage the learning of models whose predictions satisfy equality of odds for high-dimensional EHR data with sensitive attributes of more than two groups. In a setting such as ASCVD risk prediction, with a clear clinical intervention associated with the prediction, this procedure ensures that no group bears a disparate burden of mistreatment due to misclassification. However, we note that this comes at a cost of a reduction in AUC-ROC and AUC-PRC for some subgroups.

\subsection{Limitations of the Predictive Model}
While using EHR data allowed a high-capacity ASCVD risk prediction model to be trained using a large and diverse cohort, this model should not be directly compared to the PCEs for several reasons. First, the PCEs estimate ten-year ASCVD risk, whereas our model estimates risk over a period of at least a year. Furthermore, we cannot rule out the existence of biases that may lead to differential rates of selection or censoring in our cohort across age, gender, and race based subgroups, nor can we establish whether the nature of these biases differ from those present in the prospective cohort studies used to derive the PCEs.

\subsection{Moving Beyond Equality of Odds}
While we have demonstrated empirically that adversarial learning procedures are capable of encouraging a model to satisfy equality of odds, the use of this metric as a measure of fairness should be approached with caution. In the case that there is insufficient information in the training dataset to learn a high performing model for at least one group, optimizing for this criteria will upper bound the group-level model performance by the performance obtained for the least-well performing group.
In the adversarial learning setting, this reduction in performance for some groups may be offset by performance gains for groups for which the model performs poorly when trained naively. However, we observed that if such a benefit exists, it is smaller than the reduction in performance incurred for most groups.

We have not examined the relationship between the errors of the predictive model and notions of long-term utility when deploying the model clinically. To properly analyze the effect of these errors on utility requires careful causal modeling of the sequential decision-making process following ASCVD risk prediction while accounting for individual patient characteristics. We emphasize that while such a process is crucial to evaluate the long-term impact of any prediction model, it is not possible to properly identify and model that causal process with observational data in the EHR alone \cite{Kilbertus2017}. Additionally, it is unclear that satisfying fairness constraints for a single-step decision, as in ASCVD risk prediction, aligns with the goal of equitably maximizing long-term utility, as it has been shown that satisfying fairness constraints for a static decision may actually cause long-term harm in settings where an unconstrained objective would not \cite{Liu2018}, particularly if the outcome is measured with bias due to systematic censoring \cite{Kallus2018}. We find those approaches \cite{Kusner2017,Nabi2018} that establish causal notions of fairness to be promising directions for future work, as they permit sequential decision making processes to be studied under the lens of fairness at both the group and individual level.
\section{Conclusion}
Existing approaches to ASCVD risk scoring perform poorly for the population at large, with more extreme risk mis-estimates for minority populations, inadvertently exposing those groups to excess harm. We develop an ASCVD prediction model using EHR data and show that we can encourage formal notions of fairness by reducing the variability in the FPR and FNR across groups. It is not yet known to what extent algorithmic notions of fairness align with other goals, including long-term utility maximization. We hope that our results will serve as an impetus for the community at large to investigate the fairness-utility trade-off during sequential clinical decision making resulting from fairness constraints imposed on clinical risk assessments.

\section{Acknowledgements}
We would like to thank Sam Corbett-Davies, Julia Daniels, Sebastian Le Bras, and Minh Nguyen for insightful discussion and support.

This material is based upon work supported by the National Science Foundation Graduate Research Fellowship Program DGE-1656518; NLM R01 LM011369-06; NLM T15 LM007033. Any opinions, findings, and conclusions or recommendations expressed in this material are those of the authors and do not necessarily reflect the views of the funding bodies. We also thank the Widen Horizons program of the IDEX Lorraine Universit\'e d'Excellence (15-IDEX-0004) and the Snowball Associate Team funded by Inria.

\bibliographystyle{unsrt}  
\bibliography{main}

\newpage
\appendix
\counterwithin{figure}{section}
\counterwithin{table}{section}

\section{Clinical concepts used for cohort and outcome definition}
\subsection{Exclusion Criteria: Cardiovascular Disease}
The presence of any of the following ICD9CM codes prior to the index time was used to define historical cardiovascular disease: 410, 410.0, 410.00, 410.01, 410.02, 410.1, 410.10, 410.11, 410.12, 410.2, 410.20, 410.21, 410.22, 410.3, 410.30, 410.31, 410.32, 410.4, 410.40, 410.41, 410.42, 410.5, 410.50, 410.51, 410.52, 410.6, 410.60, 410.61, 410.62, 410.7, 410.70, 410.71, 410.72, 410.8, 410.80, 410.81, 410.82, 410.9, 410.90, 410.91, 410.92, 411, 411.0, 411.1, 411.8, 411.81, 411.89, 413, 413.0, 413.1, 413.9, 414, 414.0, 414.00, 414.01, 414.02, 414.03, 414.04, 414.05, 414.06, 414.07, 414.1, 414.10, 414.11, 414.12, 414.19, 414.2, 414.3, 414.4, 414.8, 414.9, 427.31, 428, 428.0, 428.1, 428.2, 428.20, 428.21, 428.22, 428.23, 428.3, 428.30, 428.31, 428.32, 428.33, 428.4, 428.40, 428.41, 428.42, 428.43, 428.9, 430, 431, 432, 432.0, 432.1, 432.9, 433, 433.0, 433.00, 433.01, 433.1, 433.10, 433.11, 433.2, 433.20, 433.21, 433.3, 433.30, 433.31, 433.8, 433.80, 433.81, 433.9, 433.90, 433.91, 434, 434.0, 434.00, 434.01, 434.1, 434.10, 434.11, 434.9, 434.90, 434.91, 436
\subsection{Exclusion Criteria: Lipid Lowering Drugs}
The presence of any of the following medications from the Antomatical Therapeutic Chemical Classification System (ATC) in the five years prior to index time was used to exclude patients on the basis of a history of prescription of lipid lowering drugs: atorvastatin (C10AA05), simvastatin (C10AA01), rosuvastatin (C10AA07), lovastatin (C10AA02), pitavastatin (C10AA08), fluvastatin (C10AA04), pravastatin (C10AA03), atorvastatin and ezetimibe (C10BA05), cerivastatin (C10AA06), atorvastatin and amlodipine (C10BX03), lovastatin and nicotinic acid (C10BA01), simvastatin and ezetimibe (C10BA02), simvastatin and fenofibrate (C10BA04).
\subsection{Outcome Definition: ASCVD}
The presence of any of the following ICD9CM codes following the index time was used to define ASCVD events (myocardial infarction and stroke): 410, 410.0, 410.00, 410.01, 410.1, 410.10, 410.11, 410.2, 410.20, 410.21, 410.3, 410.30, 410.31, 410.4, 410.40, 410.41, 410.5, 410.50, 410.51, 410.6, 410.60, 410.61, 410.7, 410.70, 410.71, 410.8, 410.80, 410.81, 410.9, 410.90, 410.91, 430, 431, 433, 433.0, 433.01, 433.1, 433.11, 433.2, 433.21, 433.3, 433.31, 433.8, 433.81, 433.9, 433.91, 434, 434.0, 434.01, 434.1, 434.11, 434.9, 434.91, 436.

Patients were also labeled as positive for ASCVD if one of the following ICD9CM codes indicative of coronary heart disease was assigned following the index time and death occurred within one year of the diagnosis: 411, 411.0, 411.1, 411.8, 411.81, 411.89, 413, 413.0, 413.1, 413.9, 414, 414.0, 414.00, 414.01, 414.02, 414.03, 414.04, 414.05, 414.06, 414.07, 414.1, 414.10, 414.11, 414.12, 414.19, 414.2, 414.3, 414.4, 414.8, 414.9

\end{document}